\documentclass[sigconf]{acmart}

\AtBeginDocument{%
  }

\setcopyright{acmlicensed}
\copyrightyear{2024}
\acmYear{2024}

\acmConference[ACM KDD'24 RelKD 2024 Workshop: submitted on 6/30/2024, accepted in July 2024]{fengshuang@gmail.com}{August 25-29}{Barcelona, Spain}
\acmISBN{978-1-4503-XXXX-X/18/06}

\renewcommand\footnotetextcopyrightpermission[1]{}
\begin{document}

\title{An Extremely Data-efficient and Generative LLM-based Reinforcement Learning Agent for Recommenders}

\author{Shuang Feng}
\authornote{Shuang Feng is the corresponding author and the technical contributor of the paper}
\email{fengshuang@gmail.com}
\affiliation{%
  \institution{Stanford University SCPD}
  \city{Palo Alto}
  \state{California}
  \country{USA}
}

\author{Grace Feng}
\email{gracefeng@ucsb.eud}
\authornote{Grace Feng assisted data processing, running experiments, and editing the paper}
\affiliation{%
  \institution{University of California Santa Barbara}
  \city{Santa Barbara}
  \state{California}
  \country{USA}
}

\renewcommand{\shortauthors}{Feng et al.}

\begin{abstract}
Recent advancements in large language models (LLMs) have enabled understanding webpage contexts, product details, and human instructions. Utilizing LLMs as the foundational architecture for either reward models or policies in reinforcement learning has gained popularity - a notable achievement is the success of InstructGPT \citep{instructGPT}. RL algorithms have been instrumental in maximizing long-term customer satisfaction and avoiding short-term, myopic goals in industrial recommender systems, which often rely on deep learning models to predict immediate clicks or purchases.

In this project, several RL methods are implemented and evaluated using the WebShop \citep{webshop} benchmark environment, data, simulator, and pre-trained model checkpoints. The goal is to train an RL agent to maximize the purchase reward given a detailed human instruction describing a desired product.

The RL agents are developed by fine-tuning a pre-trained BERT model with various objectives, learning from preferences without a reward model, and employing contemporary training techniques such as Proximal Policy Optimization (PPO) as used in InstructGPT \citep{instructGPT}, and Direct Preference Optimization (DPO) \citep{DPO}. This report also evaluates the RL agents trained using generative trajectories. Evaluations were conducted using Thompson sampling in the WebShop simulator environment.

The simulated online experiments demonstrate that DPO outperforms PPO in data efficiency and task performance, especially in success rate, using the same amount of training time. However, longer training time is necessary for fair comparison between the two. Specifically, without utilizing any image, a DPO agent achieved a 19\% success rate after approximately 3000 steps or 30 minutes of training on T4 GPUs, compared to a PPO agent, which reached a 15\% success rate after 2 hours of training. Results also indicate that agents trained on generated trajectories exhibited comparable task performance to those trained using human trajectories. This has demonstrated an example of an extremely low-cost data-efficient way of training reinforcement learning agents.  
\end{abstract}



\maketitle

\footnote{This paper was originally part of the class project for CS234 Spring 2024 in Stanford University and was submitted to KDD'24 on 6/30/2024 for RelKD Workshop. It was accepted in July 2024. See https://github.com/fengshuang-coding/KDD2024 for updates.}

\section{Introduction}
\label{Introduction}

Recent advances in Large Language Models (LLMs) have significantly enhanced research and applications in understanding human instructions on the web, processing webpage text, and grasping context. These advancements have provided valuable tools for training reinforcement learning (RL) agents to navigate web environments, particularly in e-commerce and various recommender systems such as YouTube and Netflix. Leveraging LLMs in RL agent training is relatively new but has proven successful. A notable example is InstructGPT \citep{instructGPT}, where an RL agent was trained using human preferences by fine-tuning GPT-3 models with human instructions. Combining LLMs with RL techniques enables the creation of intelligent web agents that can understand human instructions and complete tasks in web or app environments, thereby maximizing desired rewards.

Recommender systems have evolved from collaborative filtering \citep{matrixfactorization} to the recent surge in deep supervised learning, which predicts immediate user responses such as clicks \citep{deepYoutube2016, DeepRecommendersSurvey}. This approach has seen tremendous success in personalized user engagement. However, after several years in production, deep supervised learning algorithms have shown limitations, including: 1) a focus on optimizing short-term gains at the expense of long-term user satisfaction and retention, and 2) strong feedback loops caused by training data generated from these algorithms, which exacerbate these effects. Conversely, RL algorithms are designed to optimize long-term gains by learning policies that maximize long-term user satisfaction. RL agents are also well-known for their ability to perform sequential planning and make decisions based on the Markov Decision Process (MDP) properties \citep{top-k_RL_in_youtube}.

The training of RL agents for recommenders in web environments has been actively studied, with several benchmark datasets and trained agents available. For example, WikiNav \citep{task_oriented} provides a benchmark for web-based navigation RL agents. RecoGym \citep{RecoGym} offers a benchmark for RL agents in production recommendations for online advertising. Virtual-Taobao \citep{VirtualTaobao} includes a virtual online shopping environment derived from Taobao, hosting several RL algorithms for product recommendations. WebShop \citep{webshop} presents a simulated e-commerce web environment with over 1,600 human demonstrations for web shopping tasks based on human text instructions. This environment includes 1.18 million products with text and image descriptions, along with 12,087 crowd-sourced text instructions. The authors of WebShop also explored several imitation and RL agents trained using real-world human trajectories.

Previous explorations in RL for web-based recommenders are extensive. Query reformulation, as published in \citep{task_oriented}, is part of an RL problem aimed at optimizing outcomes. In this context, search engines are considered black boxes, and the RL agent (or reformulator) learns to generate queries that maximize the expected return through actions in the state space. This paper, published in 2017, predates the widespread use of BERT \citep{BERT}. The authors proposed a PRF framework, with CNN/RNN serving as the contextual learner and query generator. A more recent work proposed the concept of "learning to search" \citep{boost_agent}, where a search agent mimics the interactive process by generating interactive search queries based on previous queries and saving the best queries along the way. The authors used the T5 model with fine-tuning as a query generator to interact with the search engine iteratively, producing a set of fine-grained queries that yield better outcomes. Another related work, WebGPT \citep{webgpt}, utilizes a web interface and a search engine to train RL agents to answer questions.



\section{Related Work}
\label{Background}
The work presented in this paper is built and evaluated within the WebShop \citep{webshop} environment, a simulator of online web shopping recommender system. 

\subsection{The WebShop Environment}

WebShop is a benchmark project designed to train reinforcement learning algorithms in a large-scale, interactive, web-based environment. It includes over 12,000 crowdsourced human instructions, over 1.1 million products scraped from amazon.com. A total of 670 attributes were derived from concatenated product titles and descriptions using bi-gram representations and assigned to each product through TF-IDF scoring.

Figure 1 and Figure 2 below provide an example WebShop interface and a sequence of actions. 

\begin{figure}[h]
\caption{WebShop Environment \citep{webshop}}
\centering
\includegraphics[scale=0.45]{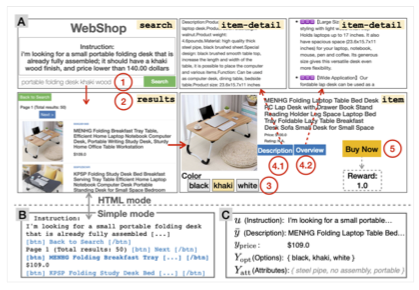}
\end{figure}

\begin{figure}[h]
\caption{WebShop Human Instructions and Human Trajectories \citep{webshop}}
\centering
\includegraphics[scale=0.45]{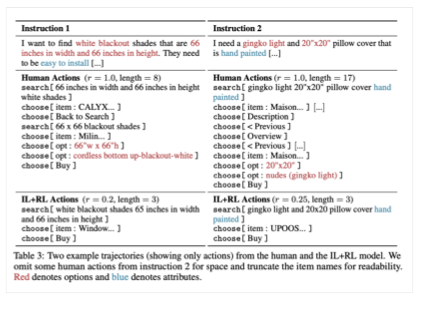}
\end{figure}

The original paper tackles the problem into two sets of reinforcement learning models for search and choice (or clicks). The search model is an imitation learning model (search-IL) mimicking human search queries from instructions. It is a human instruction and query pair fine-tuned BART \citep{bart} model in root. For choice (clicks) learning, the authors present a few reinforcement learning models to optimize choice of clicks navigating the recommender simulator to optimize the end rewards (purchase). The reward is calculated by a scoring function to quantify the relevance between a purchased product and the human instruction, based on attributes of the product. The imitation learning algorithm (choice-IL) presented by the original authors is a human trajectory fine-tuned BERT \citep{BERT} model in root. The reinforcement learning algorithm for choice iterates the imitation learned (choice-IL-RL), fine-tuned BERT model as the baseline and iterates the optimization using a mixed objectives of policy gradients and cross entropy. \\

The state space in this problem consists abstraction of four types of webpages: search page, product recommendation page, product page, and product detail page. The search page features only a search bar for entering instructions, which are used to take a search query either generated by human or a search agent, serving as input for a search engine. Actions include searching, clicking buttons, and choosing from a drop-down menu. Clicking the purchase button marks the end of a trajectory. State transitions are initiated by clicks and other actions that deterministically redirect from one webpage (state) to another. Observations, which includes state and instruction at a specific time snapshot, together form the input for the reinforcement learning agent to make subsequent actions. \\

The search engine used by the WebShop project is self-built and self-indexed offline using Pyserini \citep{Pyserini}, which is built upon the open-source Lucene search library. Product retrieval is based on BM25 between search queries and product information text. Top 50 results are shown in 5 pages ranked by BM25. \\

\subsection{Reinforcement Learning with Human Preference - RLHF}

RLHF \citep{rlhf} together with PPO \citep{PPO} were successfully used to train a few well-known GPT related product, such as instructGPT \citep{instructGPT}. RLHF leverages the Bradley-Terry model, which defines the preference using rewards of preferred and dispreferred data labeled by human labelers:\\
\begin{equation*}
    P(y_l \succ y_w|x) = \frac{exp(r(x,y_l))}{exp(r(x,y_l)) + exp(r(x,y_w))}.
\end{equation*}
RLHF objectives then can be defined similarly to entropy loss. \\

\subsection{PPO for Regularized Policy Gradient}

Proximal Policy Optimization (PPO) \cite{PPO} has been demonstrated to be effective in fine-tuning GPT models with human instructions and labeled preferences \cite{instructGPT}. PPO uses clipping or KL divergence constraints to minimize the likelihood of large updates between steps, approximately providing guarantees for monotonic improvement. This approach converges in probability to local optima and, in practice, results in more stable training outcomes. The clipped loss function for policy gradient in PPO can be expressed as:
\begin{equation}
L_{\theta_k}^{PPO} = -E_{\tau\sim \pi_k}\left[min(z_t(\theta)\hat{A}_{t}^{\pi_k},clip(z_t(\theta),1-\epsilon,1+\epsilon)\hat{A}_t^{\pi_k}\right]   
\end{equation}
, where \\
\begin{equation*}
    z_t(\theta) = \frac{\pi_\theta(a_t|o_t)}{\pi_{\theta_k}(a_t|o_t)},
\end{equation*}
\begin{equation*}
    \hat{A}_t^{\pi_k}=R_t - V^{\pi_k}(o_t).
\end{equation*}

\subsection{Learning with Human Preference - DPO}

The development of Direct Preference Optimization (DPO) \citep{DPO} is revolutionary. It eliminates the need for explicit reward functions for preferences and instead relies solely on paired preference trajectories as training data. DPO is derived from joining the Bradley-Terry objective\\
\begin{equation*}
    L_{BT}(r,D)=-E_{(x,y_w,y_l)\sim D}\left[log\ \sigma\left(r(x,y_w)-r(x,y_l)\right)\right]
\end{equation*}
, and the RLHF objective: \\
\begin{equation*}
    \smash{\displaystyle\max_{\pi}}\ E_{x\sim D,y\sim \pi}[r(x,y)]-\beta D_{KL}[\pi(y|x)||\pi_{ref}(y|x)].
\end{equation*}

The DPO loss function is:\\
\begin{equation*}
\begin{split}
    L_{DPO}(\pi_\theta,\pi_{ref}) 
    & = -E_{(x,y_w,y_l)\sim D}[log\ \sigma(\beta log\frac{\pi_\theta(y_w|x)}{\pi_{ref}(y_w|x)}\\
    & -\beta log\frac{\pi_\theta(y_l|x)}{\pi_{ref}(y_l|x)})].
\end{split}    
\end{equation*}
, where $\pi_\theta$ is the DPO policy to learn and $\pi_{ref}$ is the pre-selected reference policy. From the form of the loss function, although DPO does not need a reward model to be trained explicitly, it does require a pre-defined reference policy to iterate upon. 


\section{Approach}

This report summarizes a few efforts implementing and evaluating PPO 
vs. contrast learning using DPO using human trajectories and generated unpreferred trajectories. Then it demonstrates the contrast learning effort using all generative trajectories. \\

We branched and implemented DPO and PPO using the original WebShop code package, together with a new generative module for the self-generative learning experiments, and a Thompson sampling module to roll out online experiments and collect results. \\

For PPO training, the policy gradient objective from the original paper \citep{webshop} is modified into a PPO objective as shown in equation (1). The overall objective components, which is the total loss from policy gradient (PG), entropy loss, and imitation learning loss remain the same as in the original paper, except PG component is replaced with PPO loss. 

\subsection{Semi-generative Reinforcement Learning Using Human Trajectories}

For this project, we utilize a pre-trained imitation learning agent checkpoint as the reference policy to generate unpreferred trajectories. Preferred trajectories are obtained from human data provided by the WebShop benchmark. During training, a human trajectory is randomly sampled, including states and available actions from the log. At each state where an action decision is needed, an unpreferred action is generated using the reference policy. This unpreferred action is paired with the preferred action generated by the human. The DPO update is applied after each episode based on the human trajectory.

This approach is considered both generative and semi-self-learning. It is generative because we use a predefined unpreferred policy to generate actions for pair-wise training. It is semi-self-learning because it pairs these generated actions with previously collected human trajectories, which serve as the gold standard.\\

\subsection{Self-learning - Training with Generated Trajectories}

In classic reinforcement learning, self-play or learning through simulation plays a crucial role, particularly when data collection is costly, such as the collection of human trajectories in this problem. Self-play has proven to be effective, with the most notable example being AlphaGo \citep{alphago}.

To evaluate the idea of self-play or self-learning in navigating the WebShop recommendation systems, we generated 100 preferred trajectories using a straightforward method of sampling trajectories with perfect reward (score = 1). This sampling was done using the agent checkpoint from imitation learning provided by the authors of the WebShop paper, but with real-world human instructions.

Ideally, these sampled trajectories are pruned to eliminate looped sub-trajectories. A DPO agent is then trained from the same checkpoint used for DPO evaluation in the previous section, with 3000 steps. Task performance between these two DPO agents — one trained using semi-learning with human trajectories and the other using self-learning with generated trajectories — is compared using Thompson sampling ran in the WebShop simulator environment.



\section{Experimental Results}

\subsection{DPO vs. PPO Task Performance}

In this project, leveraging the WebShop environment and simulator, we conduct extensive simulated online experiments using Thompson sampling to analyze the performance differences across trained agents. The goal of Thompson sampling is to select the optimal action (or "arm") that minimizes overall regret. However, when sampling over a small number of steps, it may not be ideal for estimating rewards from arms that are perceived as less optimal due to insufficient exploration. To address this, we use multiple parallel runs of Thompson sampling, each with 1000 rollouts, to capture variability across runs. Careful experimental design and calculated rollouts of online experiments are necessary for accurately estimating the rewards and success rates of each agent. The aim of this project is to implement and understand the performance trends across different approaches.

The results indicate that Direct Preference Optimization (DPO) agents achieve significantly higher scores and success rates compared to Proximal Policy Optimization (PPO) agents, even though all agents start from the same imitation learning BERT model checkpoint provided by the original paper. It is important to note that all agents in this comparison are trained without image data, so the scores and success rates collected are not directly comparable to the original paper, which includes image data in training and experiments.

An interesting finding is that DPO agents trained using human trajectories perform similarly to DPO agents trained using generated trajectories, albeit with larger variance in success rate across runs. The smaller variance observed in self-learning agents can be attributed to the fact that only 100 generated trajectories were used to train the DPO self-learning agent, compared to 1200 human trajectories used for training the DPO agent with human data.

The fact that DPO agents are trained using only 3000 steps also suggests the possibility of underestimating data inefficiency or bottleneck when training over long period of times using the same set of data. When training an agent for production systems, the limited number of available trajectories can result in decreased task performance due to insufficient information learned from the limited data. In reality, collecting human data is expensive and time-consuming. This issue can be mitigated by generating preferred and unpreferred trajectories to serve as a continuous, low-cost source of training data. 

\begin{figure}[hbt!]
\caption{DPO vs. PPO --- Human Trajectories and Generated Trajectories --- Scores}
\centering
\includegraphics[scale=0.24]{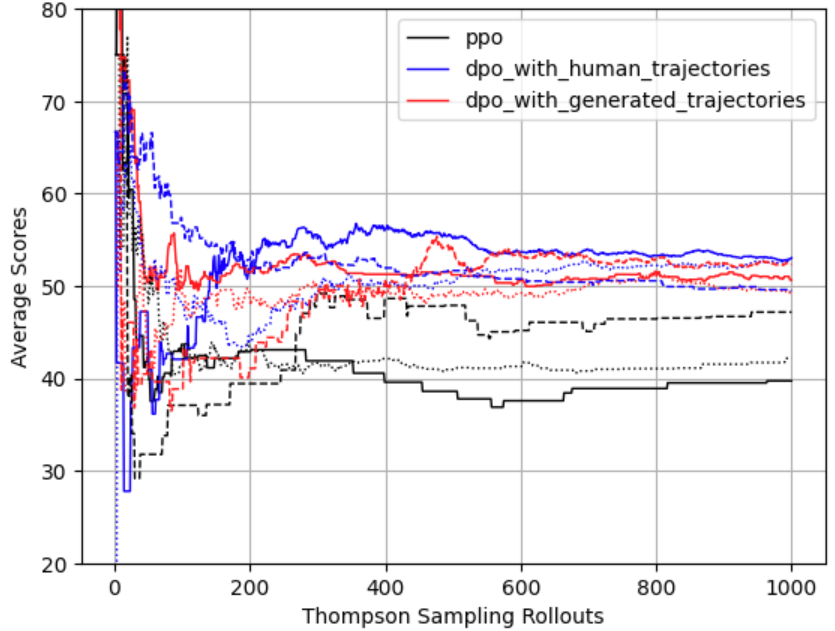}
\end{figure}

\begin{figure}[hbt!]
\caption{DPO vs. PPO --- Human Trajectories and Generated Trajectories --- Success Rate}
\centering
\includegraphics[scale=0.24]{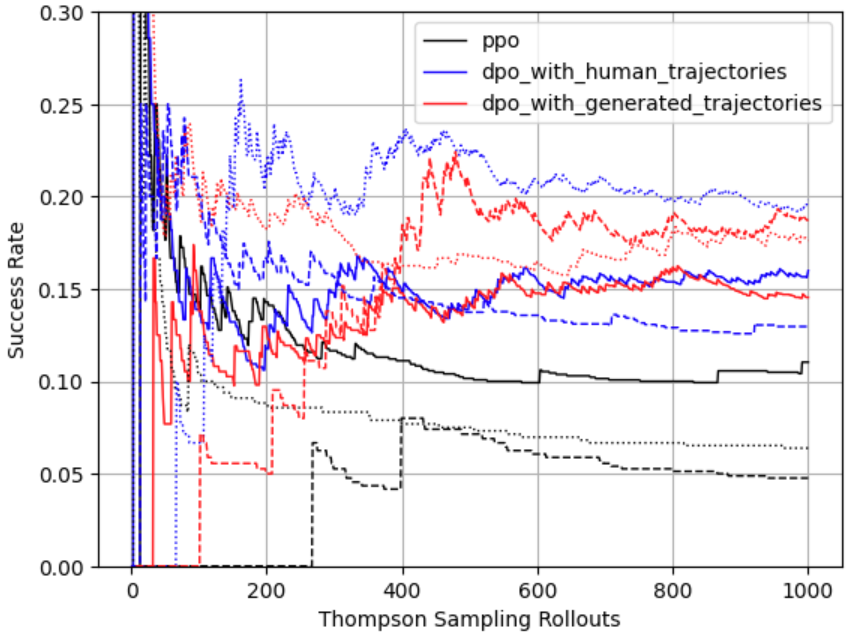}
\end{figure}

It is important to note that the results of this project are not directly comparable to those of the original paper due to two key differences: 1) no image data were used for training or experiments for any of the agents evaluated in this project, and 2) each agent was trained with minimal steps (3000) and within a timeframe of less than one hour. The purpose of this project is not to benchmark results but to investigate variations in reinforcement learning algorithms.

Training using fully generated preferences on top of the DPO agent achieved much higher scores than DPO agents using human trajectories, while the success rate remained similar (\textit{Figure 5}, \textit{Figure 6}). The magnitude of this difference needs to be justified using variance across runs, but this finding demonstrates the potential of using generative data to enhance training on top of existing agents initially trained with human trajectories.

\begin{figure}[hbt!]
\caption{Self-learning Using Generated Trajectories --- Scores}
\centering
\includegraphics[scale=0.24]{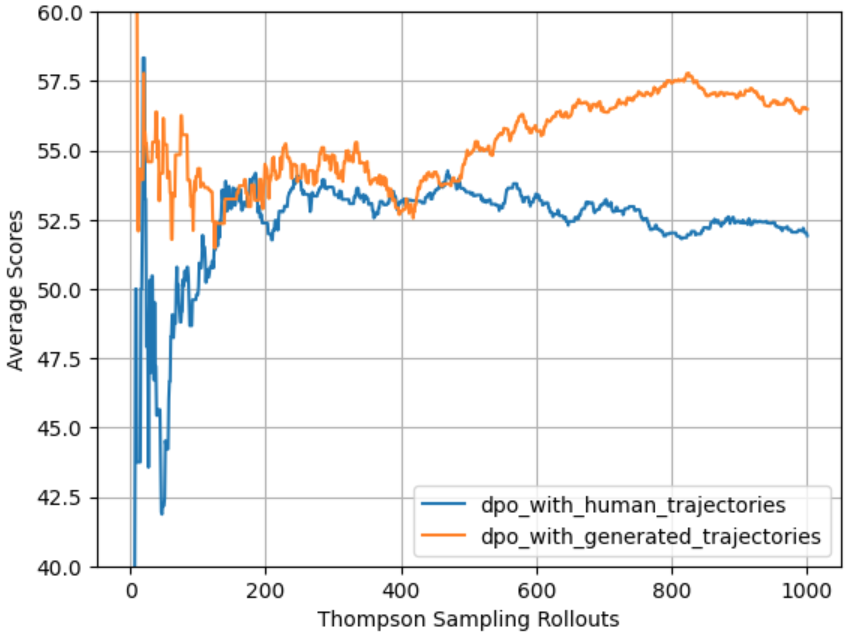}
\end{figure}

\begin{figure}[hbt!]
\caption{Self-learning Using Generated Trajectories --- Success Rate}
\centering
\includegraphics[scale=0.24]{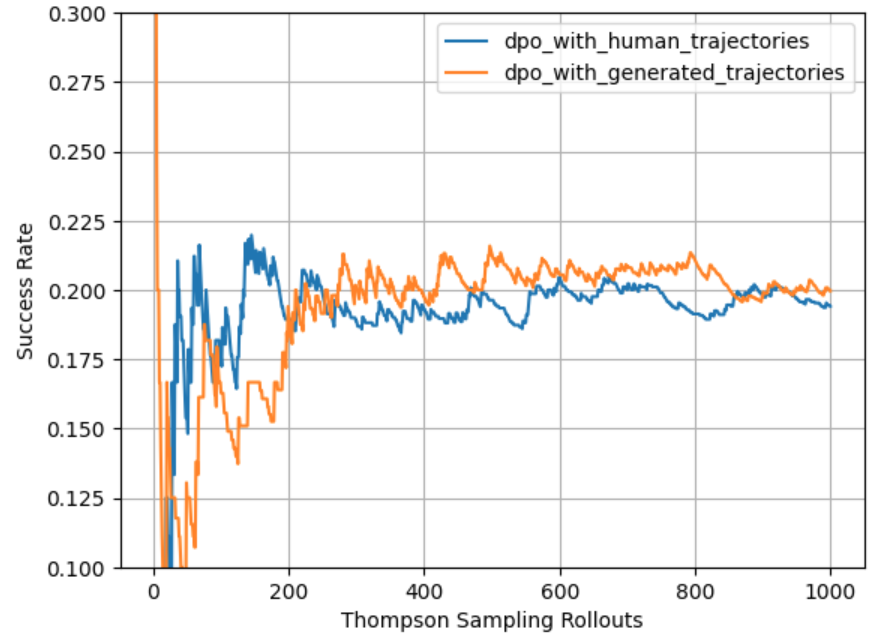}
\end{figure}

\section{Conclusion}

With very limited training time (<1 hour), Direct Preference Optimization (DPO) outperforms Proximal Policy Optimization (PPO), offering better task performance and higher success rates with less training time. However, more evaluations with longer training time are necessary to draw a conclusion. Using a DPO agent trained within one hour and without image data, we achieved a success rate of approximately 19\%. This is higher than the success rate of an RL agent trained with an RNN network (without pre-trained models for search or choice imitation learning), which has an 18\% success rate from the original paper.\\

PPO is known to be able to provide less volatile training and approximately monotonic guarantees for RL objectives. By nature, PPO's regularization and clipping of objectives prevent rapid policy changes, making it suitable for problems with smaller state and action spaces where large policy changes are not expected. However, in the context of online product recommenders, where the state-action space can expand to millions of dimensions, and rapid policy changes are essential for fast learning, PPO can need longer time to train. \\

Training DPO agents with generated trajectories has shown great potential. With only 100 generated human trajectories and the same amount of computational resources, the task performance was comparable to a DPO agent trained using 1200 human trajectories. This approach addresses data inefficiency and the high costs of human data collection. As training requires more time and data, the limited availability of human data can hinder continuous improvement. This exercise demonstrates that generated trajectories can be nearly as effective as human trajectories and can even serve as a continuous, low-cost source of training data. Additionally, generated trajectories allow exploration of successful paths not seen by humans, similar to the approach that contributed to the success of AlphaGo\citep{alphago} and AlphaZero, which were trained using self-play rather than past human games.\\

\section{Potential Usage: Using Trained Agents as a Recommender}

Using reinforcement learning agent in recommenders is not new - it is known to be used in online recommenders such as Youtube. The trained optimal policy can become an ideal ranking algorithm for recommender systems. Starting from a human instruction, the agent simulates navigating through a provided list of product following the trained optimal policy and provides a "purchased" product from each run. When using in recommenders, multiple runs of the agent provide a list of recommended product to user and the order to present in a user interface, such as on web or in an app can be rank-ordered by scores or success for each recommended product from each run of the RL agent.

\begin{acks}
To the WebShop authors Shunyu Yao, Howard Chen, John Yang and Karthik Narasimhan from Princeton University who published \citep{webshop}, which inspired this project report. \\
To Professor Chris Potts who has introduced WebShop [17] to the
author of this report, and for his excellent teaching CS224U in
Stanford University. \\
To Professor Emma Brunskill for her excellent teaching CS234 in
Stanford University. \\

\end{acks}

\bibliographystyle{ACM-Reference-Format}

\end{document}